\documentclass{article}
\usepackage[margin=1.25in]{geometry}

\usepackage{amsfonts,amsmath,amssymb,bbold,bm} % Various math commands
\usepackage{graphicx} % To include pictures
\usepackage{fancyvrb} % To include code with \VerbatimInput
\usepackage{verbatim}
\usepackage{array,multirow,rotating,siunitx}
\usepackage[ruled]{algorithm2e} % for algorithm

\usepackage{pifont}
\usepackage{amssymb}
\usepackage{hyperref}

\newif\ifcomments
\commentstrue % Change to \commentsfalse to remove all comments
\ifcomments
	\newcommand{\carlo}[1]{\textcolor{red}{[#1]}}
	\newcommand{\shuzhi}[1]{\textcolor{blue}{[#1]}}
    \newcommand{\hannah}[1]{\textcolor{magenta}{[#1]}}
    \newcommand{\shuai}[1]{\textcolor{purple}{[#1]}}
\else
	\newcommand{\carlo}[1]{}
	\newcommand{\shuzhi}[1]{}
    \newcommand{\hannah}[1]{}
    \newcommand{\shuai}[1]{}
\fi

\newcommand{\ba}{\mathbf{a}}
\newcommand{\bb}{\mathbf{b}}
\newcommand{\bc}{\mathbf{c}}
\newcommand{\bff}{\mathbf{f}}
\newcommand{\bg}{\mathbf{g}}

\newcommand{\bp}{\mathbf{p}}
\newcommand{\bq}{\mathbf{q}}

\newcommand{\bv}{\mathbf{v}}

\newcommand{\bx}{\mathbf{x}}
\newcommand{\by}{\mathbf{y}}
\newcommand{\bu}{\mathbf{u}}

\title{\textbf{Unsupervised Flow Refinement near \\ Motion Boundaries}}
\author{
    Shuzhi Yu, Hannah Kim, Shuai Yuan and Carlo Tomasi \\ \\ Duke University \\ \texttt{\{shuzhiyu,hannah,shuai,tomasi\}@cs.duke.edu}
}
\date{}

\def\etal{\emph{et al.}}
\def\ie{\emph{i.e.}}

\begin{document}
\maketitle

\begin{abstract}
Unsupervised optical flow estimators based on deep learning have attracted increasing attention due to the cost and difficulty of annotating for ground truth. Although performance measured by average End-Point Error (EPE) has improved over the years, flow estimates are still poorer along motion boundaries (MBs), where the flow is not smooth, as is typically assumed, and where features computed by neural networks are contaminated by multiple motions. To improve flow in the unsupervised settings, we design a framework that detects MBs by analyzing visual changes along boundary candidates and replaces motions close to detections with motions farther away. Our proposed algorithm detects boundaries more accurately than a baseline method with the same inputs and can improve estimates from any flow predictor without additional training.
\end{abstract}

\section{Introduction}
\label{sec:intro}
% Optical flow problem is important and unsupervised methods is practical.
Optical flow estimation is an important problem in computer vision as it enables high-level tasks such as motion segmentation~\cite{Klappstein2009, Narayana2013CoherentMS}, action recognition~\cite{Sevilla-Lara2019}, and object tracking~\cite{Vihlman_Visala_2020}. Unsupervised prediction has been attracting increasing attention~\cite{janai2018unsupervised, wang2018occlusion, meister2018unflow, liu2019selflow, arflowl2020liu, stone2021smurf} since annotating real video is both expensive and difficult~\cite{yuan2022optical}, and it is still not clear how well synthetic video can simulate real data. The state-of-the-art \emph{average} End-Point Error (EPE) of both unsupervised ~\cite{arflowl2020liu, stone2021smurf} and supervised ~\cite{PWC, raft} estimators on benchmark datasets has been decreasing ever since the introduction of Convolutional Neural Networks (CNNs) for this problem. The \emph{maximum} EPE of the estimated flow, on the other hand, is typically much larger and occurs mainly along Motion Boundaries (MBs).

\begin{figure}[ht]
    \begin{center}
        \setlength{\tabcolsep}{5pt}
        \begin{tabular}[b]{cc}
            \includegraphics[width=0.45\textwidth]{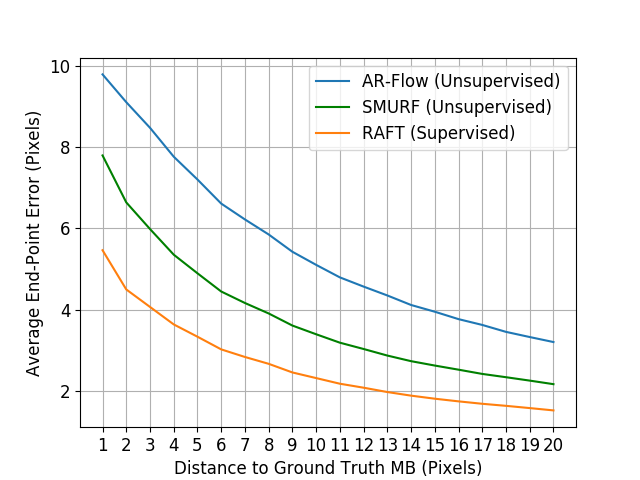} & \includegraphics[width=0.45\textwidth]{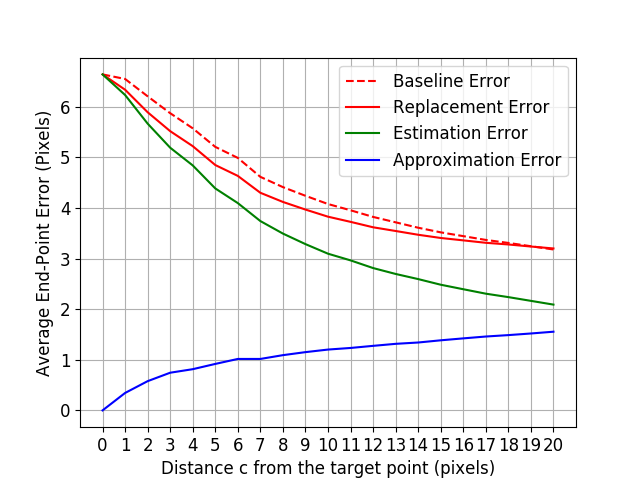}
        \end{tabular}
    \end{center}
    \caption{Left: EPE, flow estimation error $e = \|\hat{F} - F\|$ ($\hat{F}$ is the estimate, $F$ is true flow), for a top supervised flow estimator, RAFT~\cite{raft} (\textit{orange}), and two top unsupervised flow estimators, AR-Flow~\cite{arflowl2020liu} (\textit{blue}) and SMURF~\cite{stone2021smurf} (\textit{green}) versus distance to the closest true MB, averaged on all of MPI Sintel (clean). Estimates worsen near MBs. Right: Plots of the two error sources for replacement flow. Every point $\bp$ that is 2 pixels away from a true MB is a replacement target point. If $\bb$ is the closest MB point to $\bp$, let a unit vector $\bu$ point from $\bb$ to $\bp$. A point $\bq = \bp + c\bu$, for $c \in [0, 20]$, is used for this plot as long as every point in the line segment from $\bp$ to $\bq$ is within the frame and closer to $\bb$ than to any other MB point. For these $\bp, \bq$ pairs, the plot shows the estimation error $e = \|\hat{F}(\bq) - F(\bq)\|$ (\textit{green}), the approximation error $a = \|F(\bq) - F(\bp)\|$ (\textit{blue}), and the replacement error $r = \|F(\bp) - \hat{F}(\bq)\|$ (\textit{solid red}) averaged on all of MPI-Sintel (clean) and with $\hat{F}$ values from SMURF~\cite{stone2021smurf}. Flow replacement is favorable (the solid red line is under the EPE at $\bp$, \textit{dashed red}) over a wide range of values for $c$.
    %The AEPE of target points $\bp$ is not constant across different $c$ (i.e., dashed red line is not flat) because the set of target points $\bp$ used in computing the plot is different at different $c$. The target points that only appear on small $c$ are either close to image boundaries or multiple MBs, and hard to estimate flow.
    % For a target point $\bp$, suppose $c_{max}$ is the largest $c$ that satisfy the criteria. Smaller $c_{max}$ indicates $\bp$ is either close to image boundaries or multiple MBs. In either of the two cases, it is harder to accurately estimate the flow for $\bp$ and thus, the EPE increases as $c$ decreases.
    }
    \label{fig:demo_epe_vs_mbdis}
\end{figure}

% Further prove that EPEs suffer near MB, and explain why this happens.
To illustrate the performance degradation near MBs, Figure~\ref{fig:demo_epe_vs_mbdis} (left) shows the performance of top flow estimators, both unsupervised and supervised, stratified by pixel distance to the closest MB on MPI-Sintel. The EPE increases with decreasing distance from a boundary, regardless of whether supervision is available. Fundamentally, estimating motion near MBs is harder than elsewhere. First, flow is discontinuous across MBs, while typical estimators assume smoothness. Second, the features matched to find point correspondences between frames have wide receptive fields~\cite{flownet,PWC,refine_of_occ,raft} and straddle MBs when they are near them. Appearance on the two sides of a boundary typically changes in different ways from frame to frame, because of the different motions. As a consequence, feature correspondences across frames are often poor along boundaries, and this often results in poor flow estimates.

% We propose a framework on detecting MB and improving flow around them. Why MB detection and flow around them is important.
In this paper, we propose a method to detect MBs and refine optical flow near them \emph{without supervision}. Our method first detects MBs from the input flow estimated by an existing unsupervised estimator. Near these boundaries, it improves each flow estimate by replacing it with the flow a bit farther away from the MB. The possible improvement on the \emph{average} EPE over the whole image is bounded by the fact that MBs are a small fraction of any image: Only around $1\%$ of all the pixels in the dataset used in Figure~\ref{fig:demo_epe_vs_mbdis} are on true MBs. However, improvements in the \emph{maximum} EPE are important for applications that require clean MBs and accurate flow estimates near them. For example, accurate MBs would help sharpen the segmentation boundaries of moving objects in video object segmentation~\cite{caelles17osvos,maninis18osvoss} and help prevent color bleeding in the color propagation of moving objects in video editing~\cite{Meyer2018DeepVC}. In addition, recent video interpolation methods require flow estimates as input and interpolation results degrade near MBs because of poor flow estimates near them~\cite{bao18MEMC-Net,park2021ABME}.

% Why our approach is in a right direction?
There has been little work explicitly detecting MBs~\cite{unsup_mb} and improving the flow near them without supervision~\cite{monet_2021_kim, LDMB, eccv18_mb}.
%Of course, perfect flow estimation everywhere would make the detection of MBs easy. However,
%State-of-the-art flow estimators are still unable to do well enough near MBs for reliable MB detection, and we show that accurate prediction of MBs from imperfect flow estimates also helps improve flow estimates near MBs. 
% Brief introduction of MB detection
Typically, a \emph{baseline method} detects MBs by thresholding the magnitude of the flow gradient. However, results are often poor because flow estimates are both inaccurate and smooth near MBs. We show that accurate prediction of MBs from imperfect flow estimates also helps improve flow estimates near MBs. Our proposed MB detection method uses hysteresis thresholding~\cite{canny} on maps of flow gradient magnitude, image edge maps, and on novel maps we propose in this paper. These new maps identify locations in the input flow map where MBs are likely to exist, based on the observation that \emph{changes of appearance in the foreground and background on the two sides of a MB are more consistent with their own motion than with the motion on the other side.} Thus, given a point on a MB candidate, we consider two points nearby, one on each side of the boundary, and we measure how the appearance of each point would change when subjected in turn to the motions measured at either of them. If the candidate is away from a boundary, all four combinations of point and motion typically yield good matches between frames. If the candidate is on a boundary, at least two combinations often yield poorer matches.

% Brief introduction of flow method
When replacing flow values near boundaries with values farther away, we face two contrasting sources of error: The \emph{approximation error} comes from the fact that motion measured at one pixel replaces motion measured elsewhere. This error decreases as the replacement motion is taken closer to the replacement candidate. The \emph{estimation error} stems from the fact that even the replacement estimate is not exact. This error increases closer to MBs, where flow estimates degrade. Figure~\ref{fig:demo_epe_vs_mbdis} (right) shows plots of these two errors.
%which are computed based on a subset of the points used in Figure~\ref{fig:demo_epe_vs_mbdis} (left).
This trade-off explains why the replacement method cannot make improvement away from MBs, where the estimation error is flat. Even near MBs, not every point can improve through replacement. We observe that the difference in the flow values on the two sides of the MBs is a useful indicator of which points can benefit from replacement.

% Empirical result
Empirically, our MB detection method improves over the baseline methods both quantitatively and qualitatively. Our replacement algorithm improves flow at promising candidate points near MBs when compared to the state-of-the-art unsupervised flow estimators on both synthetic and real video benchmarks. We also analyze various properties of flow estimates near MBs. To the best of our knowledge, this is the first work that specifically improves and analyzes MB detection and nearby flow estimates in the unsupervised setting.

\section{Related Work}
\paragraph{Flow Estimation and Refinement}
Flow estimation has been studied for a long time~ \cite{of0,Lucas1981AnII,LDOF,deepflow,sp_of_1,Hu2017}, culminating with recent work with CNNs~\cite{flownet,ilg2017flownet2, Ranjan2017OpticalFE, PWC, liu2019selflow, refine_of_occ, liteflow1, liteflow2, liteflow3, raft}. Recently, vision transformers and attention mechanisms have also led to better results~\cite{sui2022craft, xu2022gmflow}.  Supervised CNN methods are trained on datasets like MPI Sintel~\cite{mpi} or KITTI~\cite{kitti12}. The aperture problem requires regularization, and all systems assume a smooth flow, either explicitly or implicitly and either during inference (classical methods) or during training (deep learning). State-of-the-art methods~\cite{raft} achieve good average sub-pixel accuracy, but predictions are typically at a quarter or even an eighth of the original resolution because of computation cost. Final predictions are then up-sampled, again assuming smoothness. Also, features in these systems have wide receptive fields, with the negative implications discussed earlier. As a result, the maximum EPE tends to be quite large (Figure \ref{fig:demo_epe_vs_mbdis}) even when the average EPE is small, a problem that has so far attracted little attention.

Since annotating realistic flow datasets is hard and expensive, many unsupervised CNN methods~\cite{janai2018unsupervised, wang2018occlusion, meister2018unflow, liu2019selflow, arflowl2020liu, stone2021smurf} have been proposed since the pioneer work by Yu \etal~\cite{jason2016back} and Ren \etal~\cite{ren2017unsupervised}. The top unsupervised networks often evolve from the best supervised ones. The two top unsupervised estimators used in our experiments are AR-Flow~\cite{arflowl2020liu} and SMURF~\cite{stone2021smurf}, and are based on the top supervised networks PWC-Net~\cite{PWC} and RAFT~\cite{raft} respectively. While achieving good average EPEs on benchmarks, the unsupervised flow estimators also degrade in performance near MBs.

Our method detects MBs and refines the flow near them without requiring motion labels. A related supervised method~\cite{liteflow3} refines flow by predicting a confidence map of flow and replacing low-confidence values by nearby higher-confidence values under certain conditions. However, this method ends up replacing flow away from MBs as well, which is problematic. In contrast, we replace the flow only at points near MBs, which we detect explicitly.

\paragraph{Motion Boundary Detection}
Thanks to new datasets with MB labels, recent supervised approaches use machine learning algorithms to detect MBs. LMDB~\cite{LDMB} uses structured random forests~\cite{sed} that take as inputs two consecutive images, forward and backward optical flow estimates, and image warping errors. More recent approaches use multi-task learning. Lei \etal~propose a fully convolutional Siamese network that jointly estimates both object boundaries and the motion of the pixels on them~\cite{lei2018boundary}. Ilg \etal~\cite{eccv18_mb} simultaneously estimate occlusions, depth boundaries, MBs, optical flow, disparities, motion segmentation, and scene flow (FlowNet-CSS). The state-of-the-art approach, MONets~\cite{monet_2021_kim}, jointly predicts MBs and occlusions by explicitly exploiting the relationship between the two tasks. 

These supervised methods require ground-truth flow. In contrast, our MB detection method is unsupervised. Alhersh and Stuckenschmidt~\cite{unsup_mb} work in a similar direction. They propose to use unsupervised loss to fine-tune the pre-trained flow estimators and thereby improve the detection of MBs based on the gradient of the estimated optical flow field. In comparison, we do not use any supervised pre-trained model and require no additional training. In addition, our approach detects MBs first and then improves flow estimates near them.

\section{Method}
Our method takes as input three consecutive video frames $I_1$, $I_2$, and $I_3$ and two optical flow maps $\hat{F}_{21}$ (from $I_2$ to $I_1$) and $\hat{F}_{23} \in \mathbb{R}^{h\times w\times 2}$ (from $I_2$ to $I_3$) estimated using any unsupervised flow predictor. We first detect MBs $\hat{B}_{23} \in \{0, 1\}^{h\times w}$ from frame $I_2$ to $I_3$. We then identify points $\bp$ near boundaries whose flow estimates can be potentially improved by replacing them with those at nearby points $\bq$. These replacements yield a refined map $\hat{F}^{r}_{23}\in \mathbb{R}^{h\times w\times 2}$. We now explain the components of boundary detection and flow refinement.

\subsection{Motion Boundary Detection}
% Explains the overall algorithm: 3 components. Edges and flow gradient magnitude are easy to explain.
MBs are detected by hysteresis thresholding~\cite{canny} from three feature maps, all in $\{0, 1\}^{h\times w}$: The image edge map $M_e$, the motion discrepancy map $M_{md}$, and our proposed map of invalid smooth motion $M_{ism}$. Specifically, $M_e$ is computed by the DexiNed~\cite{xsoria2020dexined} edge detector in our experiments. $M_{md}$ is obtained by thresholding the magnitude of the flow gradient. $M_{ism}$ complements $M_{md}$ by also flagging pixels where the smooth motion in the estimated flow map is unlikely to be correct, as explained below. The detector based on these maps is described in Section \ref{sec:detection}.

% Explains the feature of the sanity check of the smooth flow.
\subsubsection{Invalid smooth motion map}
Flow estimators tend to over-smooth their predictions across MBs, and the $M_{ism}$ map flags areas where the spatial smoothness of flow estimates may be unwarranted. The map is computed by analyzing two patches, one on each side of MB candidates. Let $\bb$ be a point in frame $I_2$ where the gradient $\bg$ of image brightness is nonzero and let $\bu$ be either of the two unit vectors parallel to $\bg$. MB typically align with image edges, so if $\bb$ is on a MB $\beta$, then the following two points are likely on opposite sides of $\beta$:
\begin{equation}
    \ba = \bb + \sigma\bu\;\;\;\text{and}\;\;\;\bc = \bb - \sigma\bu \;.
    \label{eq:points_on_twosides}
\end{equation}
We use $\sigma = 5$ in our experiments and order $\ba, \bc$ so that $\ba$ has the smaller EPE.

\begin{figure}[ht]
    \begin{center}
        \setlength{\tabcolsep}{15pt}
        \begin{tabular}[b]{cc}
            \includegraphics[width=0.32\textwidth]{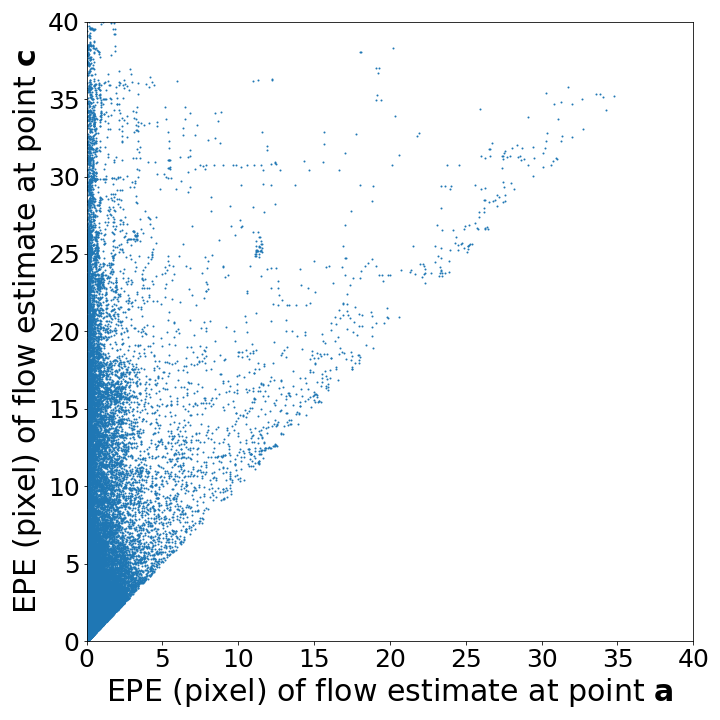} & \includegraphics[width=0.32\textwidth]{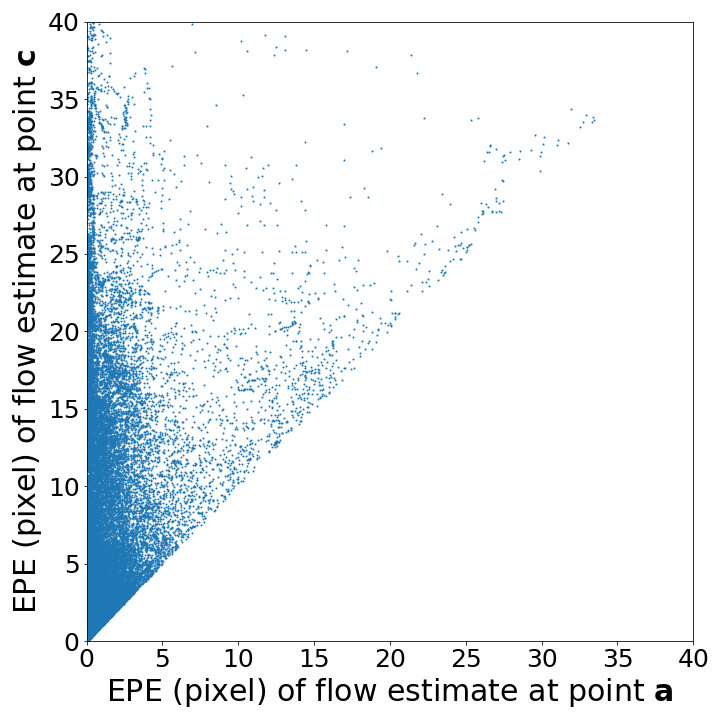}
        \end{tabular}
    \end{center}
    \caption{The scatter plots of the EPEs of point $\ba$ (x-axis) and $\bc$ (y-axis), as defined in Equation~\ref{eq:points_on_twosides}, for all the true MB points in the Sintel sequence ``alley\_2'' clean (left) and final (right) pass. Without loss of generality, $\ba$ has smaller EPE.}
    \label{fig:asymm_demo}
\end{figure}

Our proposed method for detecting the invalid smooth motion is based on the observation that the flow estimates on one side of MB are often much better than those of the other side. To illustrate, Figure~\ref{fig:asymm_demo} shows scatter plots of the EPEs of all the $\ba, \bc$ pairs for the whole Sintel sequence ``alley\_2''. The point clouds have long vertical tails in both the clean (left) and final (right) passes. For instance, in this specific sequence (clean), the accuracy of flow estimates for 54\% of the true MB points is sub-pixel at $\ba$ but not at $\bc$. Moreover, the asymmetry is 5 pixels per frame or greater in about 36\% of these asymmetric cases. The statistics are similar for the final pass. One reason for this asymmetry is that flow estimates tend to be poor on the background side of occluding MBs, where points in one frame have no match in the other.

Let $\bp$ be either $\ba$ or $\bc$. The appearance change resulting from matching point $\bp$ in frame $I_i$ to point $\bp + \bv$ by motion $\bv$ in frame $I_j$ is measured by the matching cost $c_{ij}(\bp, \bv) = -s_{ij}(\bp, \bp+\bv)$ where $s_{ij}(\cdot)$ is the Pearson correlation between the features of its arguments, \ie, $s(\bp,\bq) = \bff_{\bp}^T\bff_{\bq} / (||\bff_{\bp}||\cdot||\bff_{\bq}||) \in [-1, 1]$. In our experiments $\bff_{\bp}$ is the vector of the RGB values of a $3\times 3$ patch around $\bp$, centered by subtracting the mean patch color.

\paragraph{Matching costs under forward motion}
For simplicity, let $\hat{F}=\hat{F}_{23}$. Without knowing which side the EPE is smaller, if 
we use $\hat{F}(\ba)$ or $\hat{F}(\bc)$ to match either $\ba$ or $\bc$ in frame $I_2$ to a point in frame $I_3$, we get four cost measurements $m_{\ba\ba}$, $m_{\ba\bc}$, $m_{\bc\ba}$, $m_{\bc\bc}$, where $m_{\ba\bc} = c(\ba, \hat{F}(\bc))$ and so forth.% (see Figure~\ref{fig:demo_algo} (a)).

%\begin{figure}[ht]
%    \begin{center}
%        \setlength{\tabcolsep}{20pt}
%        \begin{tabular}[b]{cc}
%            \includegraphics[width=0.5\textwidth]{images/BMVC2022/BMVC_twosides_demo.pdf} & \includegraphics[width=0.3\textwidth]{images/BMVC2022/BMVC_repalg_demo.pdf} \\
%            (a) & (b)
%        \end{tabular}
%    \end{center}
%    \caption{(a) The four matching costs that are used in checking whether the smooth motion across a point is valid. The blue and red arrows respectively represent the estimated motion of the blue and red patches. (b) Notation for the replacement algorithm. Only the estimated flow of one side of the boundary, in set $P$, may be replaced by $\bq$. The line through $\bq$, $\bb$, $\bq'$ is perpendicular to the boundary at $\bb$, and parallel to unit vector $\bu$.}
%    \label{fig:demo_algo}
%\end{figure}

\textbf{Away from MBs} and with no occlusion, the estimated flow is often accurate. Since flow is smooth there, we typically have $m_{\ba\ba} - m_{\bc\ba} \leq \theta_{ism}$ for some pre-defined small threshold $\theta_{ism}$. Similarly, $m_{\ba\bc} - m_{\bc\bc} \leq \theta_{ism}$. However, occlusions may also occur away from MBs in the presence of large motions~\cite{monet_2021_kim}. In that case, the inequalities above are likely to hold with \emph{backward motion}, as explained later.

\textbf{Near MBs}, and where the EPE is asymmetric as explained earlier, let $\ba$ be the point with the smaller EPE, without loss of generality. We assume that the matching cost of $\bc$ is large following the motion of $\ba$, \ie~$m_{\bc\ba}$ is large. %Although the neighborhood close to $\bc$ can be textureless, a motion, very different from the true motion, would move $\bc$ out of the neighborhood of its true location in the next frame.
On the other hand, $m_{\ba\ba}$ is likely small. Similar considerations follow when $\ba$ and $\bc$ are switched, and we accordingly define the map
\begin{equation}
    M_{ism}(\bb) = \max\{m_{\ba\bc}-m_{\bc\bc},\;\; m_{\bc\ba}-m_{\ba\ba}\} > \theta_{ism}\;.
\end{equation}
There will be false negatives if the flow estimates are inaccurate on both sides of $\bb$, and false positives if there is a large matching cost away from MBs. Using backward motion and hysteresis thresholding, as explained next, will mitigate these false predictions.

\paragraph{Adding backward motion}
Suppose the motion is smooth between frames 1 and 3. An additional frame $I_1$ is helpful to the detection of the MB because an occluding MB forward in time is a dis-occluding MB in reverse time, and flow near dis-occlusions are often more accurate than near occlusions. We therefore redefine the cost as follows:
\begin{equation}
    \tilde{m}_{\bx\by} = \min\{c_{23}(\bx, \hat{F}_{23}(\by)), c_{21}(\bx, \hat{F}_{21}(\by))\}\;.
\end{equation}

% One reason for this is that motions at $\bb$, $\bc$ are different in this case, and using the wrong motion often leads to poor matches.

% Detailed overall detection algorithm.
\subsubsection{Detection algorithm}\label{sec:detection}
We use a notion similar to hysteresis thresholding~\cite{canny} to combine the edge map $M_e$, motion discrepancy map $M_{md}$, and invalid smooth motion map $M_{ism}$. Specifically, points for which $M_{md}$ is true are classified as \emph{strong MBs}. A point is classified as \emph{weak} MB where $M_{md}$ is false and both $M_e$ and $M_{ism}$ are true. Other points are classified as non-motion-boundary points. A weak MB point becomes a strong one if it is spatially connected with some strong MB points. The final strong MB points are the final predictions from the detector.

\subsection{Flow Replacement}
\label{sec:replacement_algorithm}
%Not all the points near MBs may have good substitutes for their flow because either or both of the estimation and approximation errors can be large, as discussed in Section~\ref{sec:intro}. We observe that points whose motion is very different from, and smaller than, that on the other side of a MB are likely to be improved through replacement. This is because the flow estimates at those points are often negatively affected by the large motion on the other side, and points $\bq$ on the same side but away from the boundaries are likely to have accurate flow predictions, since existing flow estimators tend to achieve better accuracy on smaller motions. Furthermore, the absolute approximation error tends to be smaller for small motions. In contrast, although the flow of the points on the larger-motion side and close to MBs are also often affected badly by the boundary, points $\bq'$ away from MBs on that side are likely to have inaccurate flow estimates as well, since predictors do more poorly on large motion. Coupled with the fact that the absolute approximation error also tends to be large with larger motions, this issue makes replacements less effective on points on the larger-motion side.

From the analysis above, replacing flow at $\bp$ with flow at $\bq$ near a MB is most effective where the improvement from a smaller estimation error at $\bq$ trumps the increasing approximation error as $\bp$ and $\bq$ are taken farther apart. This turns out to occur most often for points $\bp$ where true flow is small and true flow opposite the MB is large. At these points $\bp$, the approximation error is smaller because the flow is smaller and therefore typically changes less rapidly. The estimation error is smaller as well, because flow predictors do better on smaller motions. 

% To estimate differences in motion across a MB candidate we need to identify a point on each side of the candidate where the motion field can be estimated reliably. As shown in Figure~\ref{fig:demo_epe_vs_mbdis} (left), the error of the estimated flow decreases for points away from the MBs. Assuming smooth motion on each side of a MBs, the estimated motion of a reliable point is much closer to the flow of the point 1 pixel further away from the boundary than the flow of the point 1 pixel away from the boundary. Suppose $\bq = \bb + d\bu$ is the first reliable point on one side of a MB, relative to the MB point $\bb$. Then, $d$ is the distance of the first reliable point from a MB, and it is the smallest number such that
% \begin{equation}
%     \|\hat{F}(\bq) - \hat{F}(\bq + \bu)\| < \tau\|\hat{F}(\bq) - \hat{F}(\bb + \bu)\|\;,
%     \label{eq:find_q}
% \end{equation}
% where $\tau$ controls how strong the smoothness assumption, measured by the left-hand side of the constraint, is relative to the difference in flow between the reliable point $\bq$ and the point $\bb+\bu$. Smaller $\tau$ indicates a stronger assumption on the smooth motion, and would generally result in larger $d$.

To identify these points, a method is needed to detect points on the two sides of a MB candidate where flow can be estimated well. Since flow is smooth on either side of a MB, most of the variation in flow with the distance $d = \|\bq - \bb\|$ of $\bq$ from the MB is caused by interference from the flow across the MB.
Figure~\ref{fig:demo_epe_vs_mbdis} (left) shows that this interference tends to drop and saturate as the distance $d$ from the MB increases.
Thus, flow estimates tend to change more rapidly for smaller values of $d$ than for larger ones. Formally, let $\hat{f}(d) = \hat{F}(\bb + d\bu)$ for brevity. Then, the change $\delta(d, d+1) = \|\hat{f}(d) - \hat{f}(d+1)\|$ resulting from a one-pixel increment from $d$ to $d+1$ drops and saturates as well. We measure this drop relative to the overall change in flow from distance $1$ to distance $d$, that is, relative to $\delta(1, d) = \|\hat{f}(1) - \hat{f}(d)\|$ and define the \emph{smallest safe distance} $d^*$ to be the smallest distance for which the ratio between $\delta(d, d+1)$ and $\delta(1, d)$ drops below some threshold $\tau\in(0, 1)$:
\begin{equation}
    d^*={\min}\left\{d\;\middle\vert\;\frac{\|\hat{f}(d) - \hat{f}(d+1)\|}{\|\hat{f}(1) - \hat{f}(d)\|} < \tau\right\}\;.
    \label{eq:find_q}
\end{equation}
We use $\tau = 0.2$ in all our experiments.
Let now $\bq = \bb + d^*\bu$ be the first safe point on one side of the MB, and similarly define a first safe point $\bq'$ on the other side.
Then, we define the set $P$ containing all the replacement candidates near $\bb$ as follows:
\begin{equation}
P = \left\{\bp = \bb + d\bu\;\middle|\;\underbrace{0 < d < d^*}_{\text{close to MB}}\;\;\&\;\;
\underbrace{\|\hat{F}(\bq)\| < \|\hat{F}(\bq')\|}_{\text{side with smaller flow}}\;\;\&\;\;
\underbrace{\|\hat{F}(\bq) - \hat{F}(\bq')\|\geq \alpha\|\hat{F}(\bq)\|}_{\text{large difference across MB}} \right\}\;,
\label{eq:def_set_P}
\end{equation}
%Specifically, and with reference to Figure \ref{fig:demo_algo} (b), flow estimates are replaced only in the area $P$ of points that are within $d^*$ pixels of a MB prediction. The points in set $P$ has smaller motion than those on the other side of the MB, and the difference is controlled by $\alpha$. Larger $\alpha$ indicates larger difference in the motion of the two sides.
We use $\alpha=0.2$ in our experiments. The refined flow is
\begin{equation}
        \hat{F}^r(\bp) =
            \hat{F}(\bq) \;\;\; \text{if $\bp\in P$} \;\;\; \text{and}\;\;\;
        \hat{F}^r(\bp) =
            \hat{F}(\bp) \;\;\; \text{otherwise.}
\end{equation}

\section{Empirical Results}

\subsection{Experiment Settings} \label{exp_setting}
\textbf{Datasets:}
\textbf{MPI-Sintel}~\cite{mpi} dataset provides the optical flow labels for each frame of 23 high resolution synthetic sequences of 20 to 50 frames each in its training set. Fast motion and large occluded areas make this dataset challenging. We follow LDMB~\cite{LDMB} to compute the ground-truth MB labels. \textbf{KITTI-2015}~\cite{Menze2015ISA, Menze2018JPRS} is a realistic dataset that is commonly used as a benchmark in flow estimation. However, its training set, consisting of 200 image pairs, only provides sparse ground truth optical flow, and thus accurate true MBs cannot be inferred. We use these two training sets to demonstrate the improvement of our algorithm on the detection of MBs and the flow estimates near them.

\textbf{Performance evaluation:}
We evaluate estimated MBs by $F_1$-score. We follow the literature~\cite{LDMB,eccv18_mb} and use the BSDS benchmark~\cite{bsds} to compute MB detection performance. A prediction is a true positive as long as it is within 0.75\% of the image diagonal away from a true MB~\cite{eccv18_mb}. Optical flow is evaluated by End-Point Error (EPE).

\textbf{Flow estimators:}
All the pre-trained flow estimators are trained without ground truth labels, and are provided by their authors~\cite{stone2021smurf,arflowl2020liu}. 
% The pre-trained SMURF and AR-Flow models used in the KITTI experiments are trained on the KITTI-2015 training set; the SMURF model used in Sintel experiments is trained on Sintel test set and the AR-Flow is trained on the training set. We keep the default parameters of the classical flow estimator LDOF~\cite{LDOF} in our experiments.

\textbf{Hyper-parameters:}
The threshold for the motion discrepancy map is set to be 1 for Sintel and 3 for KITTI. For MB detection, we set the threshold $\theta_{ism}$ for the maps of invalid smooth motion as 0.2 for Sintel dataset and 0.6 for KITTI.

\subsection{Motion Boundary Detection}
In Table~\ref{tab:exp_results_mb}, we compare our method with the baseline method over different input flows to show our method's robustness. The performance is evaluated on both the clean and final passes of the MPI Sintel training set, following the literature~\cite{monet_2021_kim}. 

% The flow estimators include two top unsupervised CNN based flow estimators, SMURF~\cite{stone2021smurf} and AR-Flow~\cite{arflowl2020liu}, and one top classical flow estimator, LDOF~\cite{LDOF}, where SMURF achieves the lowest AEPEs (although the model has not been trained on the training set), and LDOF achieves the highest AEPEs. 

The table shows that the performance of both the baseline method and ours improves with better input flow, except that the baseline method does better with LDOF than with AR-Flow on Sintel clean. Our method consistently outperforms the baseline method across all three flow estimators and on both passes. The improvement ranges from 5.97\% (70.3 to 74.5) on Sintel clean with SMURF to 21.09\% (53.1 to 64.3) on Sintel clean with AR-Flow. The largest improvements on both passes are with AR-Flow. On one hand, the good performance by the baseline method with SMURF limits the margin for improvement. On the other hand, the estimated flow needs to be accurate at least on one side of a MB for good detection, and thus the inaccurate input flow from LDOF limit the improvement margin as well.

\begin{figure}[ht]
    \begin{center}
        \setlength{\tabcolsep}{5pt}
        \begin{tabular}[b]{ccccc}
            GT MB & Baseline & Edge Map & Map $M_{ism}$ & Ours \\
            \includegraphics[width=0.17\textwidth]{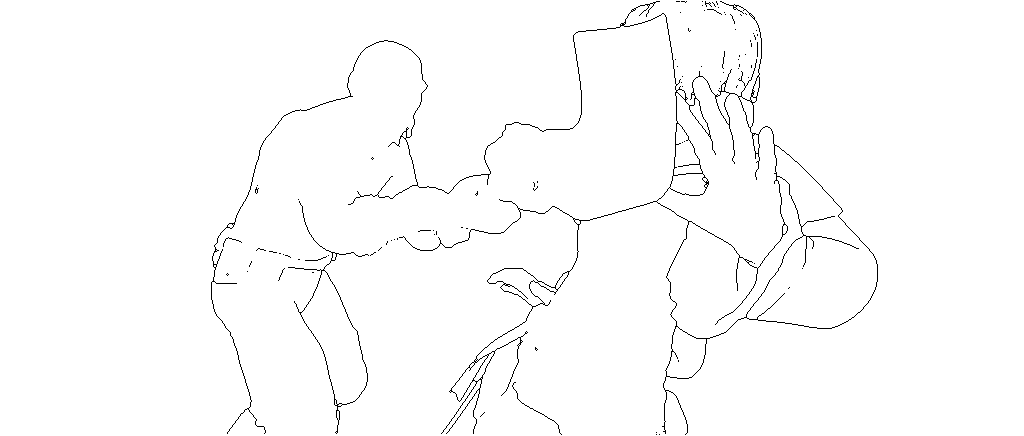} &
            \includegraphics[width=0.17\textwidth]{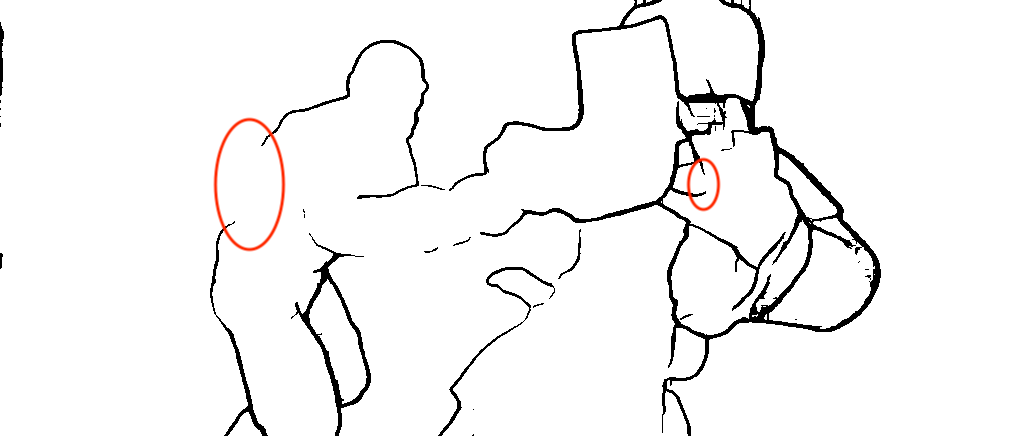} &
            \includegraphics[width=0.17\textwidth]{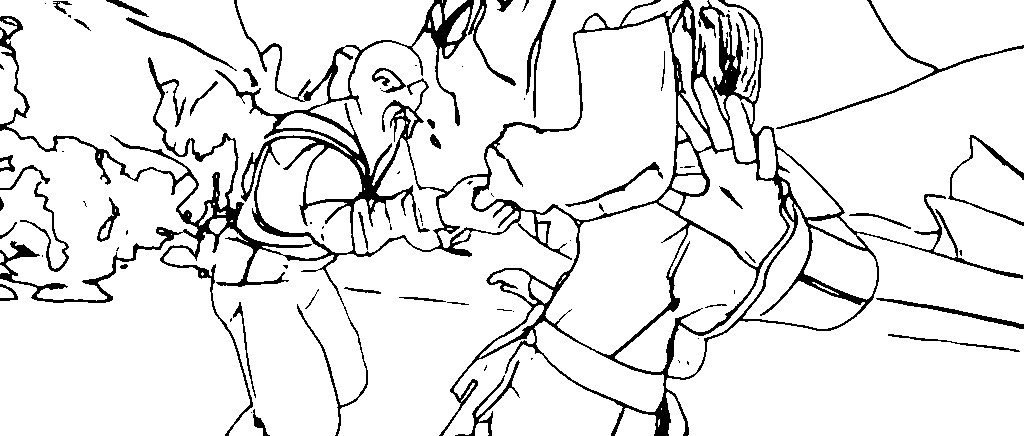} &
            \includegraphics[width=0.17\textwidth]{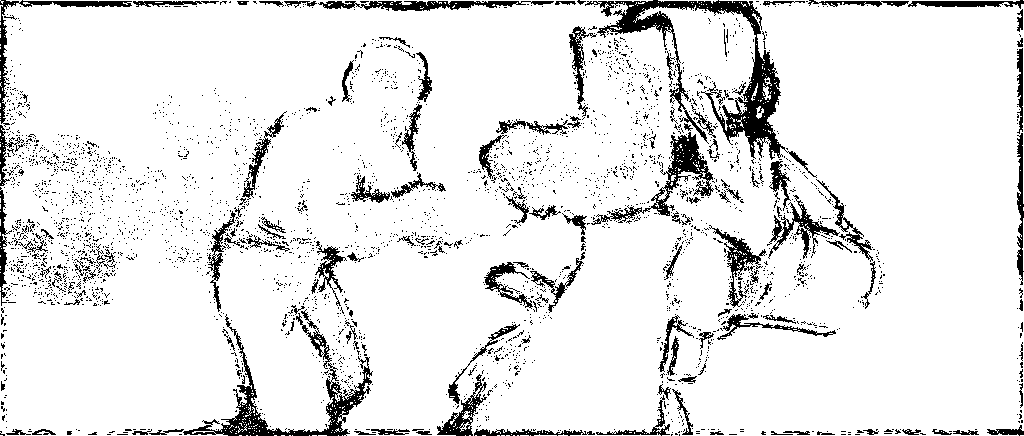} &
            \includegraphics[width=0.17\textwidth]{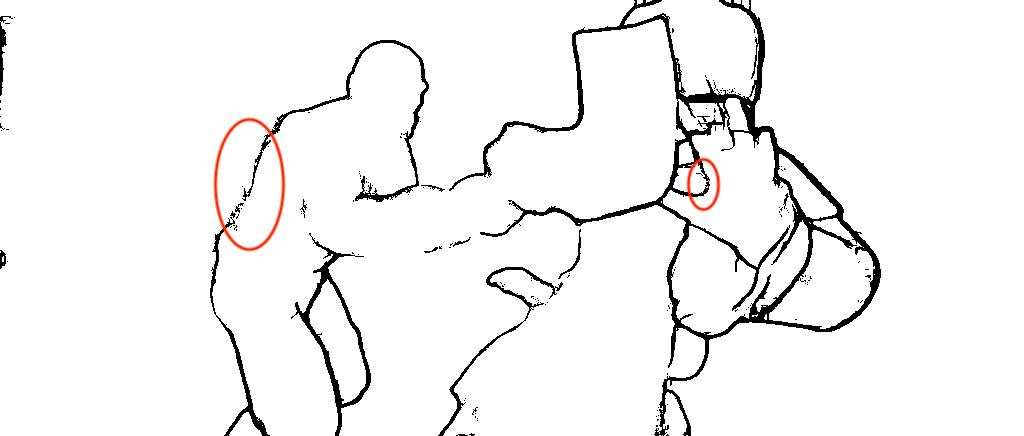} \\ \hline
            \includegraphics[width=0.17\textwidth]{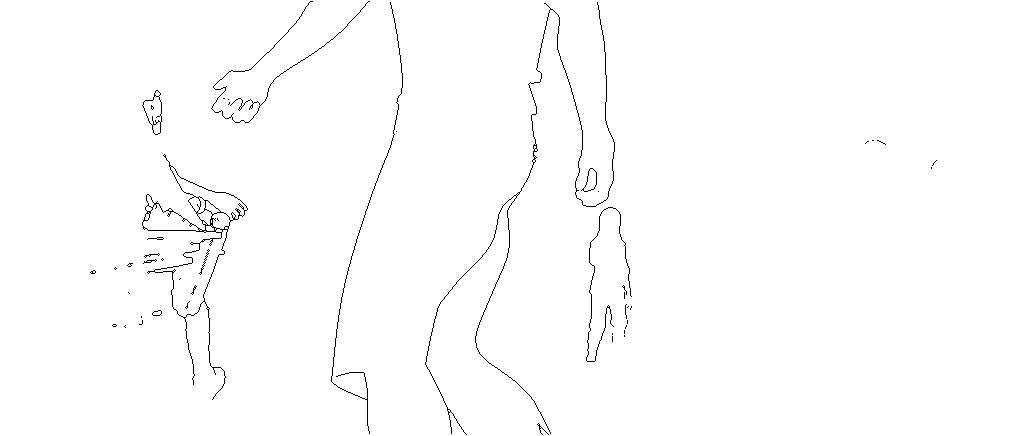} &
            \includegraphics[width=0.17\textwidth]{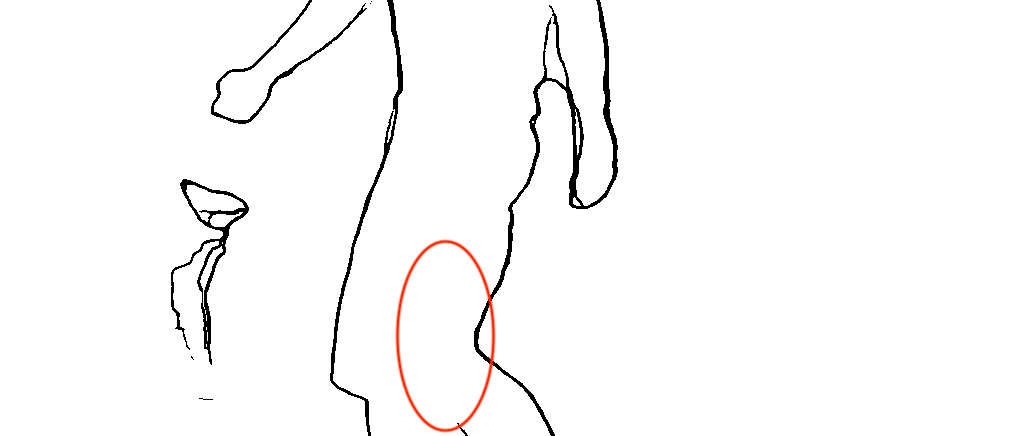} &
            \includegraphics[width=0.17\textwidth]{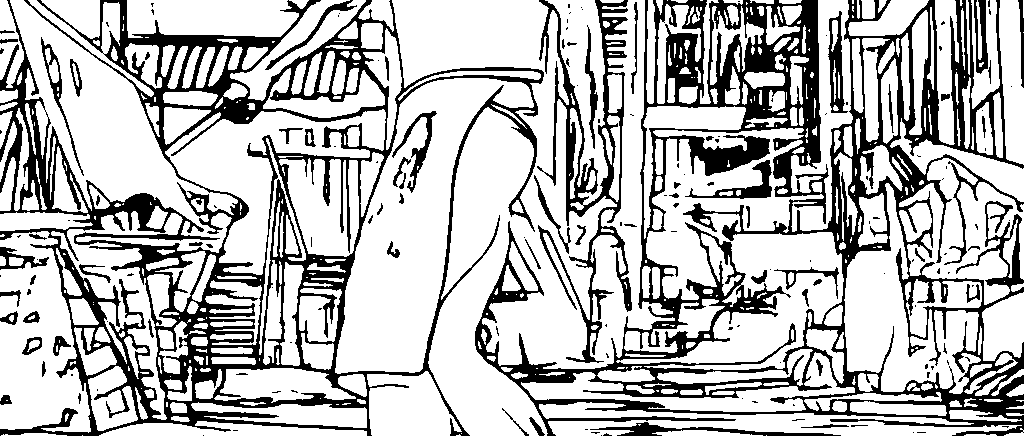} &
            \includegraphics[width=0.17\textwidth]{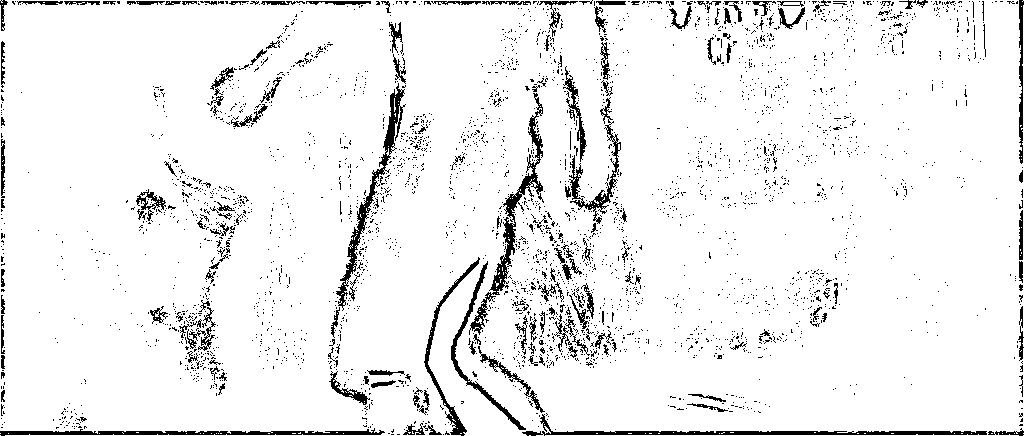} &
            \includegraphics[width=0.17\textwidth]{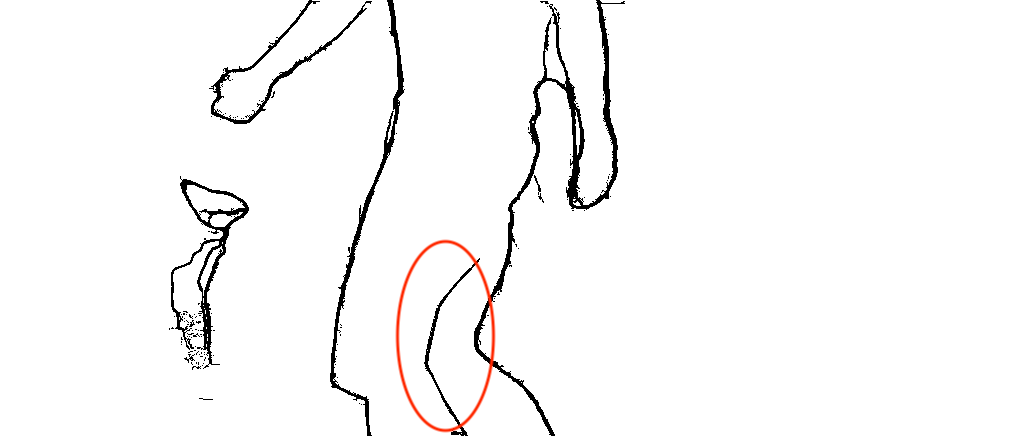}
        \end{tabular}
    \end{center}
    \caption{MB detection samples with our method and the baseline on Sintel (clean) with SMURF input. Main differences are highlighted by red ovals.}
    \label{fig:sintel_mb_example}
\end{figure}

Figure~\ref{fig:sintel_mb_example} shows two MB detection examples on Sintel (clean). Our method detects some MBs missed by the baseline method (red ovals in columns 2, 5) thanks to the invalid-smooth-motion maps, even if these are noisy (fourth column).

\begin{table}[ht]
    \begin{center}
    \begin{tabular}{cc|ccc|ccc|ccc}
    \cline{3-11}
     && \multicolumn{3}{c|}{SMURF} & \multicolumn{3}{c|}{AR-Flow} & \multicolumn{3}{c}{LDOF} \\ \cline{3-11}
     && Flow & BL & Ours & Flow & BL & Ours & Flow & BL & Ours \\
     && (EPE) & (F1) & (F1) & (EPE) & (F1) & (F1) & (EPE) & (F1) & (F1) \\ \hline
    \multirow{2}{*}{\rotatebox[origin=c]{90}{Sintel}} & Clean & 2.01 & 70.3 & \textbf{74.5} & 2.79 & 53.1 & \textbf{64.3} & 4.18 & 54.8 & \textbf{59.2} \\
    & Final & 2.87 & 63.5 & \textbf{67.4} & 3.73 & 48.5 & \textbf{57.1} & 6.25 & 46.7 & \textbf{51.2} \\\hline
    \end{tabular}
    \end{center}
    \caption{$F_1$-score for \textbf{MB estimation} with different input flow estimates, compared with the baseline method (BL). SMURF~\cite{stone2021smurf} and AR-Flow~\cite{arflowl2020liu} are two top unsupervised flow estimators, and LDOF~\cite{LDOF} is a top classical flow estimator.}
    \label{tab:exp_results_mb}
\end{table}

\textbf{Map ablation study:}
%The key to our proposed MB detection algorithm is the hysteresis thresholding using flow gradient magnitude $M_{md}$, edge map $M_e$, and our proposed map $M_{ism}$ of invalid smooth motion.
Maps $M_e$ and $M_{ism}$ (columns 3, 4 in Figure~\ref{fig:sintel_mb_example}) complement each other: The edge map localizes geometry well but does not convey motion information. The map $M_{ism}$ contains motion information but is noisy. Table~\ref{tab:ablation_two_maps_on_mb} shows that using either map alone on top of $M_{md}$ actually worsens performance. Using both of them with hysteresis improves over $M_{md}$ in both passes of Sintel.

\setlength{\tabcolsep}{7pt}
\begin{table}[ht]
    \begin{center}
    \begin{tabular}{c|c|c|c|c}
    \cline{2-5}
    & Baseline (Map $M_{md}$) & +Map $M_e$ & +Map $M_{ism}$ & Ours (+$M_e$+$M_{ism}$) \\ \hline
    Clean & 70.3 & 39.7 & 49.7 & \textbf{74.5} \\
    Final & 63.5 & 42.5 & 54.2 & \textbf{67.4} \\ \hline
    \end{tabular}
    \end{center}
    \caption{Performance of MB detection ($F_1$) of our proposed hysteresis scheme with different map combinations.}
    \label{tab:ablation_two_maps_on_mb}
\end{table}

\subsection{Flow Replacement}
%After detecting MBs, our next step is to identify both the point candidates whose flow can be improved through replacement, \ie set $P$, and replacement candidates, \ie point $\bq$. In
Table~\ref{tab:exp_results_flow} shows the effects of our flow refinement over the flow from SMURF (state-of-the-art) on points in the replacement set $P$. We consistently improve estimates over all datasets.

\begin{figure}[ht]
    \begin{center}
        \begin{tabular}[b]{cc|ccc}
            & & GT flow & Input flow & Refined flow \\
            \raisebox{8mm}{\rotatebox[origin=c]{90}{Sintel}} & \includegraphics[width=0.12\textwidth]{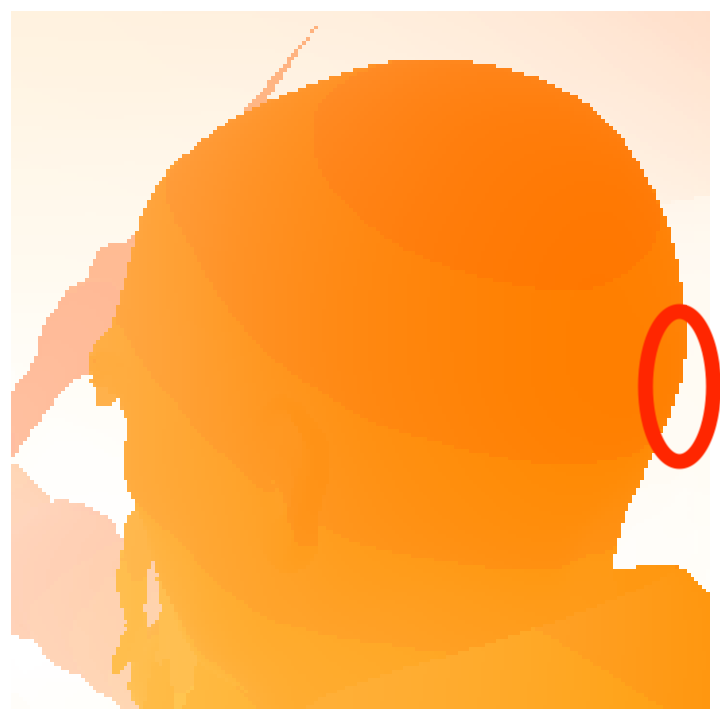} &
            \includegraphics[width=0.25\textwidth]{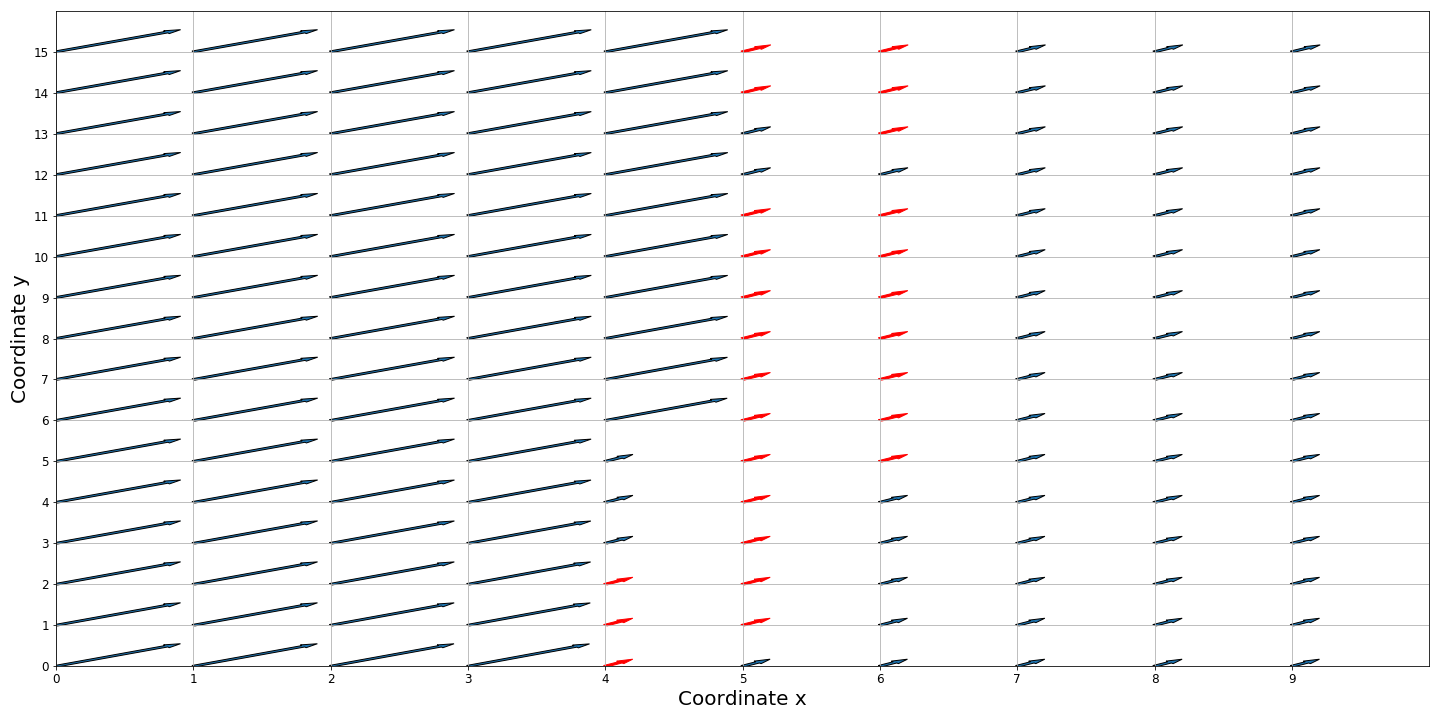} & 
            \includegraphics[width=0.25\textwidth]{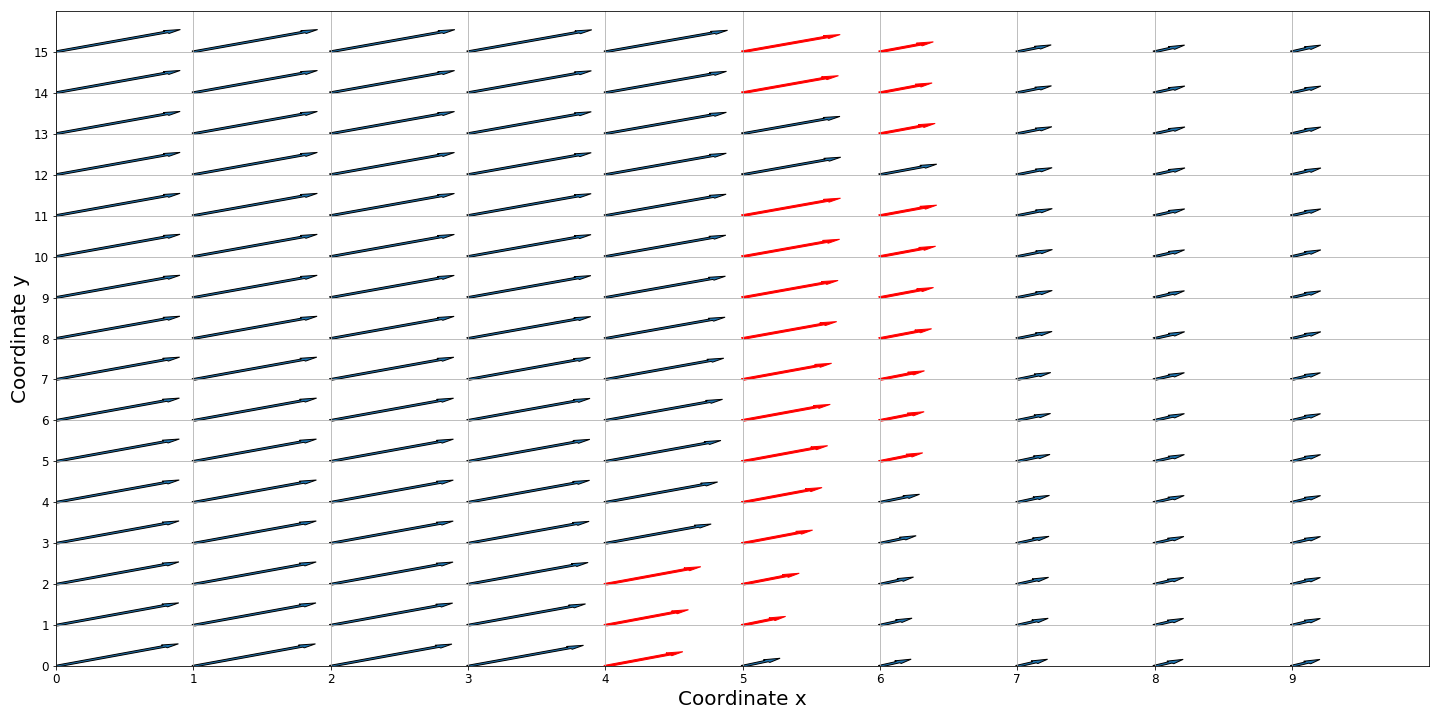} & 
            \includegraphics[width=0.25\textwidth]{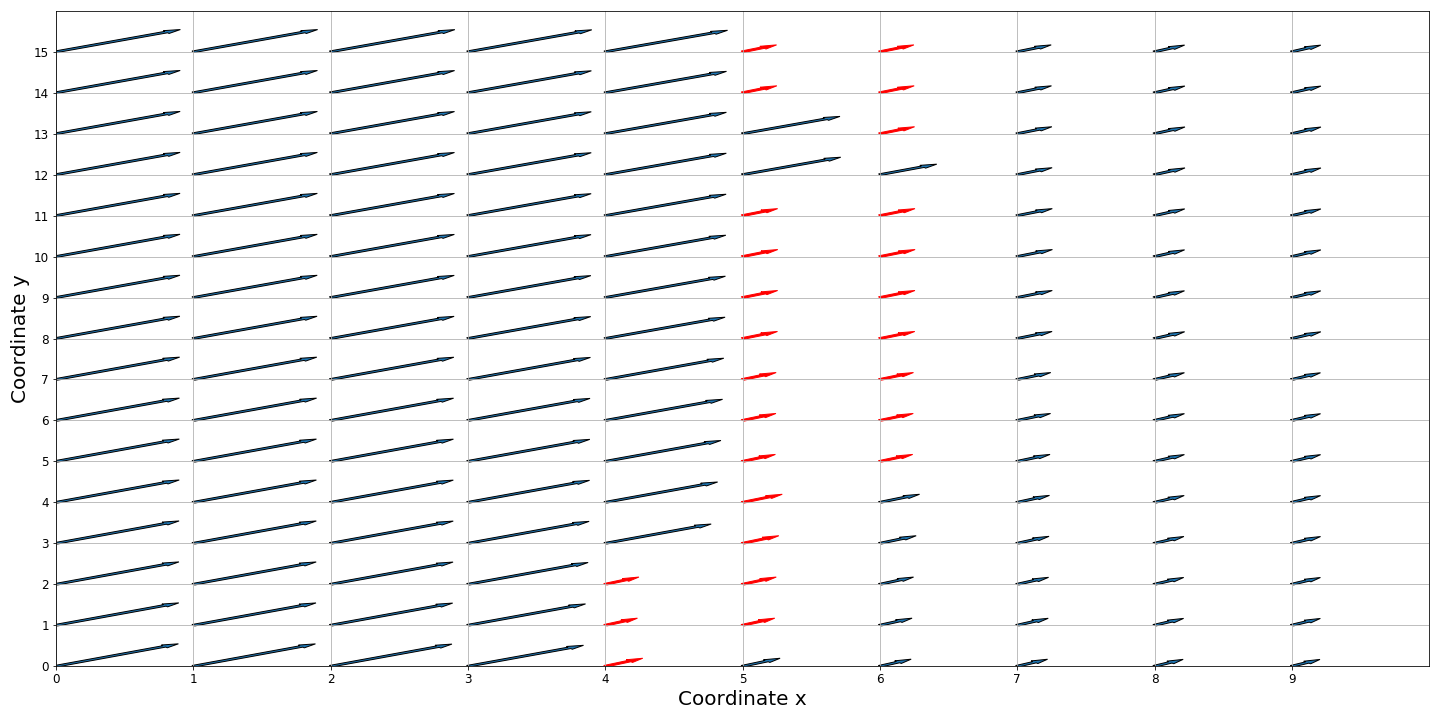} \\ \hline
            \raisebox{8mm}{\rotatebox[origin=c]{90}{KITTI}} & \includegraphics[width=0.12\textwidth]{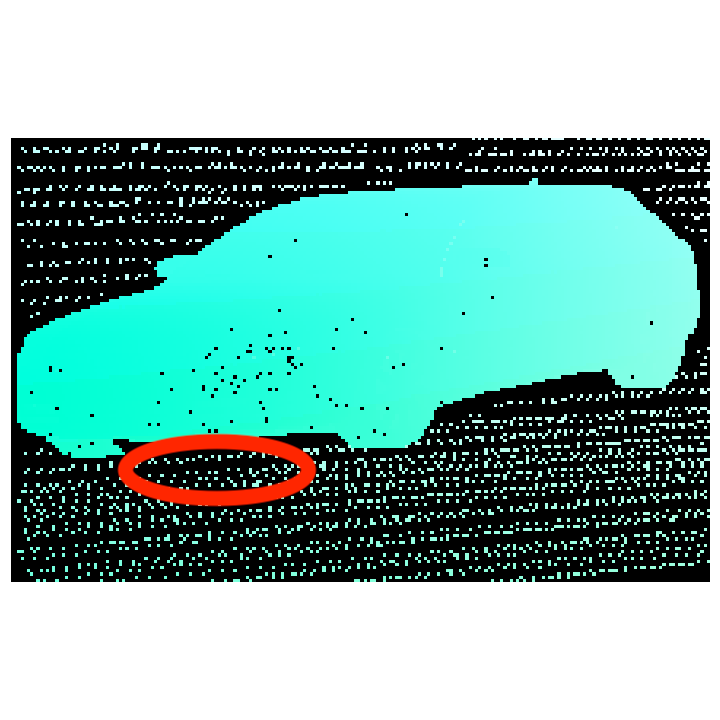} &
            \includegraphics[width=0.25\textwidth]{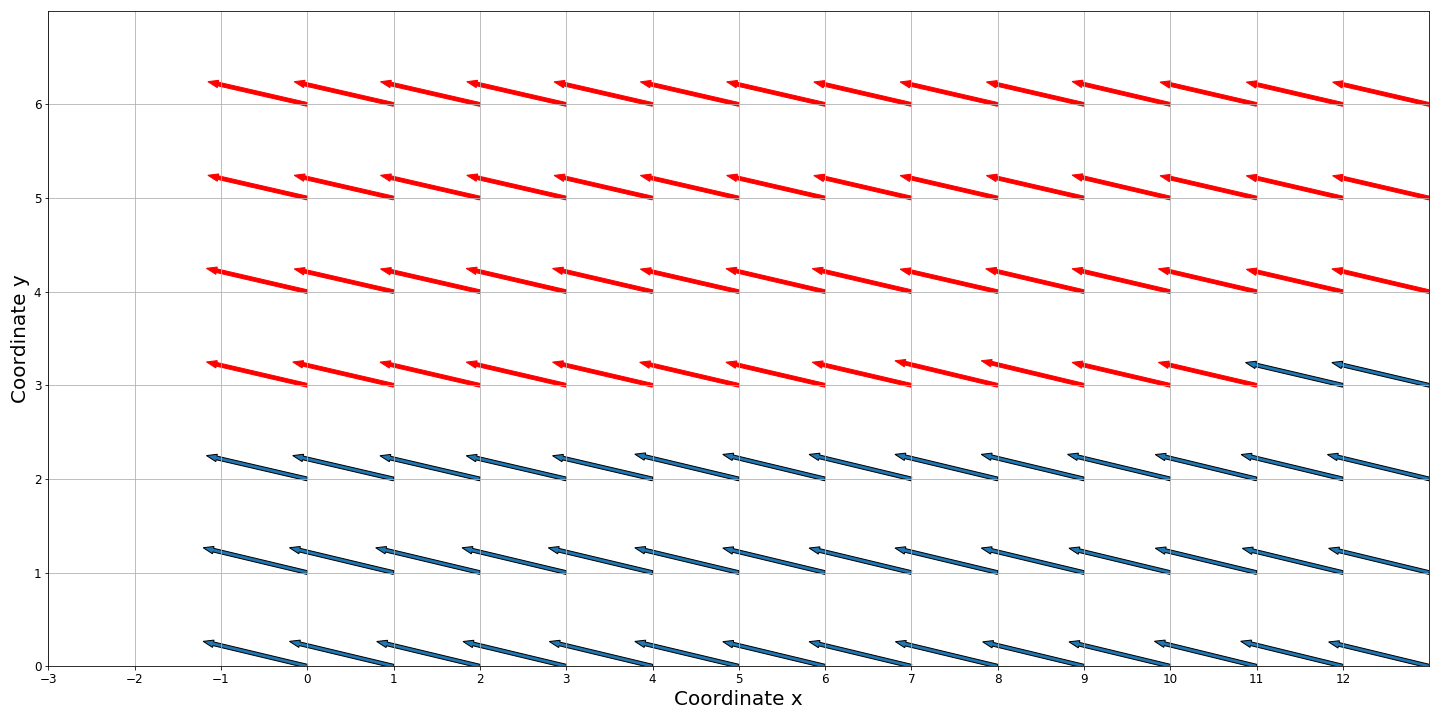} & 
            \includegraphics[width=0.25\textwidth]{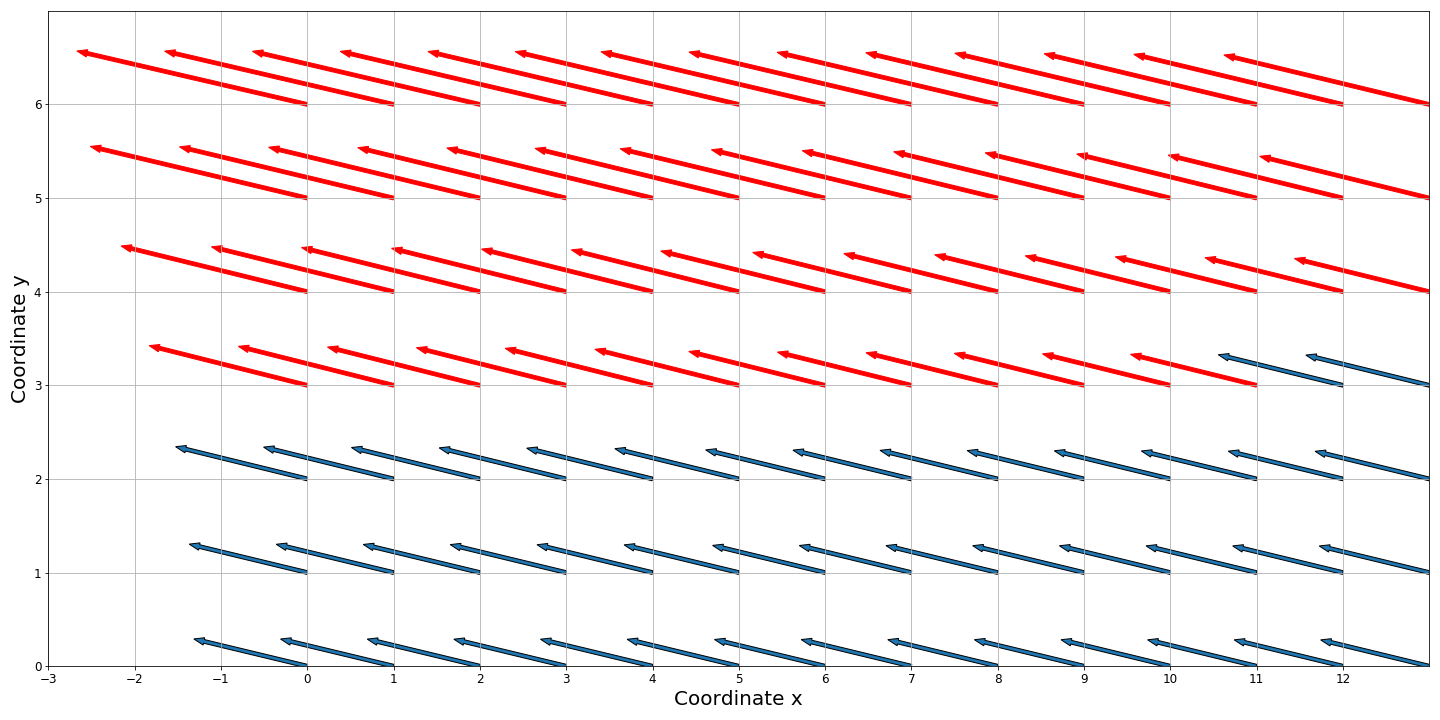} & 
            \includegraphics[width=0.25\textwidth]{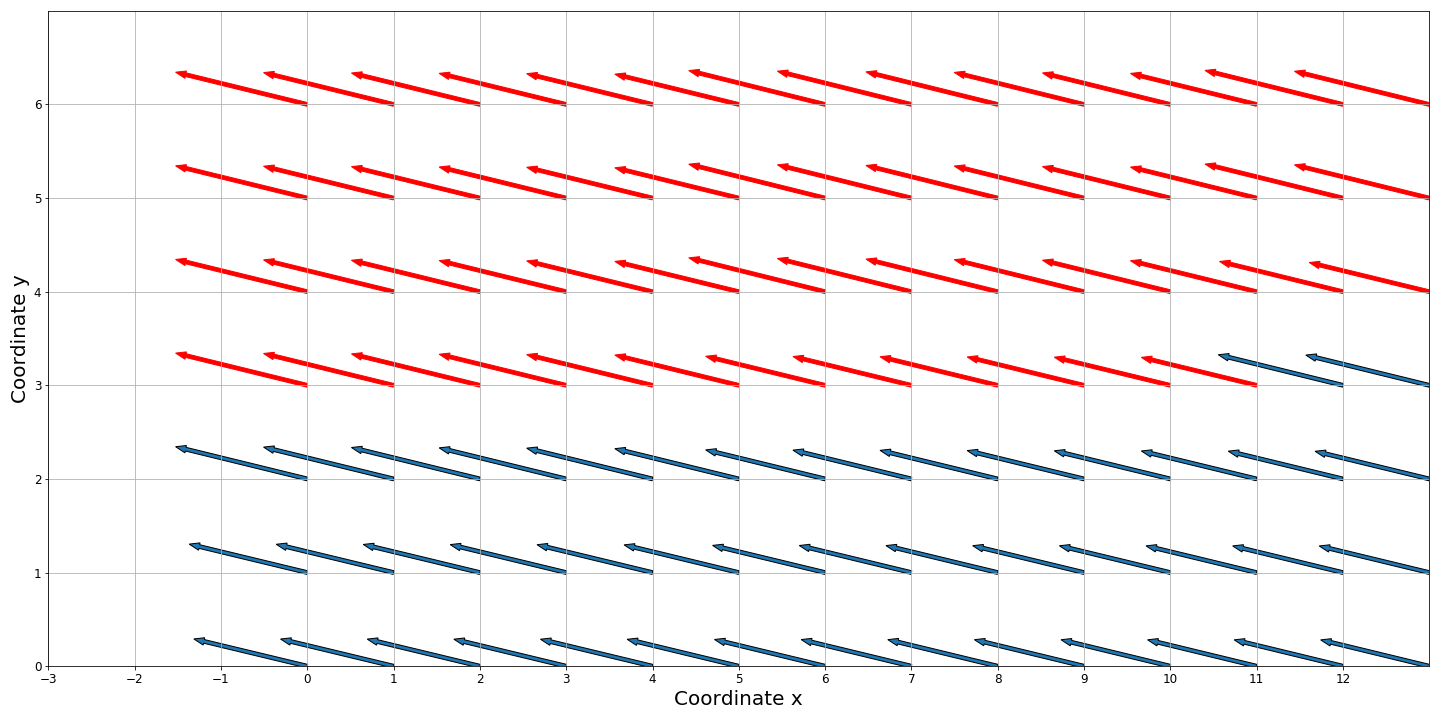} \\
            & (a) & (b) & (c) & (d)
            
        \end{tabular}
    \end{center}
    \caption{Two examples of flow (quiver plots with flow down-scaled by 70) before and after replacement on set $P$ (red vectors). Input flow is from SMURF~\cite{stone2021smurf}. On these two patches, replacement decreases the AEPE from 24.62 to 3.40 pixels per frame
    %(86.19\% improvement)
    for the Sintel example and from 33.23 to 16.87 pixels per frame
    %(49.23\% improvement)
    for the KITTI example.}
    \label{fig:qualitative_study_flow}
\end{figure}

Figure~\ref{fig:qualitative_study_flow} shows before/after flow quiver plots in two examples. Red vectors are in $P$. In the Sintel example, the set $P$ is in the background and close to the true MB. In the KITTI example, the set $P$ appears to be moved from the true MB (KITTI has no ground-truth MB labels). However, flow estimates in $P$ are still perturbed by the running car with larger motion. Replacement improves flow estimates in both cases.

% Number of true MB points in Sintel is 4432
\setlength{\tabcolsep}{4pt}
\begin{table}[ht]
    \begin{center}
    \begin{tabular}{c|c|c|c|c|c|c}
    \hline
    \multirow{2}{*}{Input Flow} & \multirow{2}{*}{Dataset} & Input Flow & \multicolumn{4}{c}{Replaced Points} \\ \cline{4-7}
    & & AEPE & \% of MB points & Init AEPE & Our AEPE & $\downarrow$ \\\hline
    \multirow{3}{*}{SMURF} & Clean & 2.03 & 61.28 & 5.47 & $\mathbf{5.17}$ & 5.48\% \\
    & Final & 2.90 & 39.98 & 5.71 & $\mathbf{4.72}$ & 17.34\% \\ 
    & KITTI & 1.94 & - & 15.35 & $\mathbf{14.69}$ & 4.30\% \\\hline
    % \multirow{3}{*}{AR-Flow} & Clean & 2.82 & 43.86 & 8.84 & $\mathbf{7.15}$ & 19.12\% \\
    % & Final & 3.78 & 22.97 & 8.43 & $\mathbf{6.77}$ & 19.69\% \\ 
    % & KITTI & 2.85 & - & 19.14 & $\mathbf{18.53}$ & 3.19\% \\ \hline
    % \multirow{3}{*}{LDOF} & Clean & 4.18 & 51.02 & 12.81 & $\mathbf{10.84}$ & 15.38\% \\
    % & Final & 6.25 & 33.24 & 13.68 & $\mathbf{11.28}$ & 17.54\% \\ 
    % & KITTI & 19.63 & - & 44.95 & $\mathbf{43.76}$ & 2.65\% \\ 
    \end{tabular}
    \end{center}
    \caption{SMURF~\cite{stone2021smurf} average EPE and average EPE improvement with our replacement method near our estimated MBs. About 1\% of all MPI Sintel pixels are true MB points. This information is unknown for KITTI, which has sparse ground truth flow.}
    \label{tab:exp_results_flow}
\end{table}

% \setlength{\tabcolsep}{4pt}
% \begin{table}[ht]
%     \begin{center}
%     \begin{tabular}{c|c|c|c|c|c|c|c|c}
%     \hline
%     \multirow{2}{*}{Input Flow} & \multirow{2}{*}{Dataset} & Input Flow & \multicolumn{2}{c|}{d0-10} & \multicolumn{2}{c|}{d10-60} & \multicolumn{2}{c}{d60-140} \\ \cline{4-9}
%     & & AEPE & Input & Ours & Input & Ours & Input & Ours \\ \hline
%     \multirow{2}{*}{SMURF} & Clean & 3.425 & 3.105 & 3.101 & 1.281 & 1.281 & 0.727 & 0.727  \\
%     & Final & 4.773 & 4.236 & 4.223 & 1.894 & 1.894 & 1.334 & 1.334 \\ 
%     \end{tabular}
%     \end{center}
%     \caption{End-point Error of existing flow estimators and the performance improvement with our proposed replacement algorithm using the MBs estimated by our proposed MB estimator (with the corresponding flow estimates as input). The performance is evaluated on the Sintel Test dataset.}
%     \label{tab:exp_results_sintel_test}
% \end{table}

\textbf{Impact of MB quality on replaced flow:}
%The success of our replacement method depends on the accurate detection of MBs. 
% The estimated flow far away from MBs is hard to improve through the replacement scheme. The estimation errors of those points are often similar but the approximation errors would cause the replaced flow worse than the original flow. 
Table~\ref{tab:mb_quality_on_flow} (top half) shows that better MBs lead to better flow replacement on Sintel (clean). This pattern is less clear on the final pass (bottom half), where improvements are comparably large regardless of the quality of the MBs. This is likely because the estimated flow is generally worse on Sintel final, especially near MBs, so there is more room for improvement by replacement.

\setlength{\tabcolsep}{4pt}
\begin{table}[ht]
    \begin{center}
    \begin{tabular}{c|c|c|c|c|c}
    \hline
    \multicolumn{2}{c|}{Dataset} & Ours (LDOF) & Ours (AR-Flow) & Ours (SMURF) & GT \\ \hline
    \multirow{2}{*}{Clean} & MB (F1) & 59.2 & 64.3 & 74.5 &  100.0 \\ 
    & EPE $\downarrow$ (\%) & 0.02 & 0.75 & 5.48 & 7.72  \\ \hline
    \multirow{2}{*}{Final} & MB (F1) & 51.2 & 57.1 & 67.4 & 100.0 \\ 
    & EPE $\downarrow$ (\%) & 17.51 & 22.50 & 17.34 & 20.94\\\hline
    \end{tabular}
    \end{center}
    \caption{Impact of MB quality (by F1) on the performance of flow replacement algorithm on Sintel clean and Sintel final. The flow performance is evaluated on the replacement set $P$ and the input flow is from SMURF~\cite{stone2021smurf}.}
    \label{tab:mb_quality_on_flow}
\end{table}

\section{Conclusion}
We propose a method that both detects MBs and improves flow estimates near them. The method is plug-and-play and requires no supervision. Fundamentally, it exploits the fact that it may be fruitful to replace a flow vector near a MB with one taken from a pixel that is farther away. This is useful because the error in taking the flow vector from the wrong point is on average smaller than the error caused by proximity to a MB. Figure \ref{fig:demo_epe_vs_mbdis} (right) shows that this balance is favorable on average, and our method exploits that margin fully. Of course, the Figure also shows that the benefit is bounded, and our analysis elucidates the trade-offs. Future work will address methods to tackle the flow estimation error near MBs directly.
\clearpage

\bibliographystyle{bmvc2k}
\bibliography{bib_paper}

\begin{thebibliography}{10}\itemsep=-1pt

\bibitem{unsup_mb}
T.~Alhersh and H.~Stuckenschmidt.
\newblock Unsupervised fine-tuning of optical flow for better motion boundary
  estimation.
\newblock In {\em International Joint Conference on Computer Vision, Imaging
  and Computer Graphics (VISIGRAPP)}, 2019.

\bibitem{bao18MEMC-Net}
W.~Bao, W.-S. Lai, X.~Zhang, Z.~Gao, and M.-H. Yang.
\newblock Memc-net: Motion estimation and motion compensation driven neural
  network for video interpolation and enhancement.
\newblock {\em IEEE Transactions on Pattern Analysis and Machine Intelligence},
  2018.

\bibitem{LDOF}
T.~Brox and J.~Malik.
\newblock Large displacement optical flow: descriptor matching in variational
  motion estimation.
\newblock {\em IEEE Transactions on Pattern Analysis and Machine Intelligence},
  33(3):500--513, 2011.

\bibitem{mpi}
D.~J. Butler, J.~Wulff, G.~B. Stanley, and M.~J. Black.
\newblock A naturalistic open source movie for optical flow evaluation.
\newblock In {A. Fitzgibbon et al. (Eds.)}, editor, {\em European Conference on
  Computer Vision}, Part IV, LNCS 7577, pages 611--625. Springer-Verlag, Oct.
  2012.

\bibitem{caelles17osvos}
S.~Caelles, K.-K. Maninis, J.~Pont-Tuset, L.~Leal-Taix\'e, D.~Cremers, and
  L.~{Van Gool}.
\newblock One-shot video object segmentation.
\newblock In {\em Computer Vision and Pattern Recognition (CVPR)}, 2017.

\bibitem{canny}
J.~F. Canny.
\newblock A computational approach to edge detection.
\newblock {\em IEEE Transactions on Pattern Analysis and Machine Intelligence},
  PAMI-8:679--698, 1986.

\bibitem{sp_of_1}
H.-S. Chang and Y.-C.~F. Wang.
\newblock Superpixel-based large displacement optical flow.
\newblock In {\em 2013 IEEE International Conference on Image Processing},
  pages 3835--3839. IEEE, 2013.

\bibitem{sed}
P.~Dollár and L.~Zitnick.
\newblock Structured forests for fast edge detection.
\newblock In {\em IEEE International Conference on Computer Vision}, 2013.

\bibitem{flownet}
A.~Dosovitskiy, P.~Fischer, E.~Ilg, P.~H{\"a}usser, C.~Haz{\i}rba{\c{s}},
  V.~Golkov, P.~v.d. Smagt, D.~Cremers, and T.~Brox.
\newblock Flownet: Learning optical flow with convolutional networks.
\newblock In {\em IEEE International Conference on Computer Vision}, 2015.

\bibitem{kitti12}
A.~Geiger, P.~Lenz, and R.~Urtasun.
\newblock Are we ready for autonomous driving? the kitti vision benchmark
  suite.
\newblock In {\em Conference on Computer Vision and Pattern Recognition}, 2012.

\bibitem{of0}
B.~K. Horn and B.~G. Schunck.
\newblock Determining optical flow.
\newblock {\em Artificial Intelligence}, 1981.

\bibitem{Hu2017}
Y.~Hu, Y.~Li, and R.~Song.
\newblock Robust interpolation of correspondences for large displacement
  optical flow.
\newblock In {\em The IEEE Conference on Computer Vision and Pattern
  Recognition}, 2017.

\bibitem{liteflow1}
T.~Hui, X.~Tang, and C.~Loy.
\newblock Liteflownet: A lightweight convolutional neural network for optical
  flow estimation.
\newblock In {\em 2018 IEEE/CVF Conference on Computer Vision and Pattern
  Recognition (CVPR)}, pages 8981--8989, Los Alamitos, CA, USA, jun 2018. IEEE
  Computer Society.

\bibitem{liteflow3}
T.-W. Hui and C.~C. Loy.
\newblock Liteflownet3: Resolving correspondence ambiguity for more accurate
  optical flow estimation.
\newblock In A.~Vedaldi, H.~Bischof, T.~Brox, and J.-M. Frahm, editors, {\em
  IEEE International Conference on Computer Vision}, pages 169--184, Cham,
  2020. Springer International Publishing.

\bibitem{liteflow2}
T.-W. Hui, X.~Tang, and C.~C. Loy.
\newblock {A Lightweight Optical Flow CNN - Revisiting Data Fidelity and
  Regularization}.
\newblock {\em {IEEE Transactions on Pattern Analysis and Machine
  Intelligence}}, 2020.

\bibitem{refine_of_occ}
J.~Hur and S.~Roth.
\newblock Iterative residual refinement for joint optical flow and occlusion
  estimation.
\newblock In {\em IEEE Conference on Computer Vision and Pattern Recognition
  (CVPR)}, pages 5747--5756, Long Beach, CA, USA, 2019.

\bibitem{ilg2017flownet2}
E.~Ilg, N.~Mayer, T.~Saikia, M.~Keuper, A.~Dosovitskiy, and T.~Brox.
\newblock Flownet 2.0: Evolution of optical flow estimation with deep networks.
\newblock In {\em Proceedings of the IEEE conference on computer vision and
  pattern recognition}, pages 2462--2470, 2017.

\bibitem{eccv18_mb}
E.~Ilg, T.~Saikia, M.~Keuper, and T.~Brox.
\newblock Occlusions, motion and depth boundaries with a generic network for
  disparity, optical flow or scene flow estimation.
\newblock In {\em The European Conference on Computer Vision}, September 2018.

\bibitem{janai2018unsupervised}
J.~Janai, F.~Guney, A.~Ranjan, M.~Black, and A.~Geiger.
\newblock Unsupervised learning of multi-frame optical flow with occlusions.
\newblock In {\em Proceedings of the European Conference on Computer Vision},
  pages 690--706, 2018.

\bibitem{jason2016back}
J.~Y. Jason, A.~W. Harley, and K.~G. Derpanis.
\newblock Back to basics: Unsupervised learning of optical flow via brightness
  constancy and motion smoothness.
\newblock In {\em Proceedings of the European Conference on Computer Vision},
  pages 3--10. Springer, 2016.

\bibitem{monet_2021_kim}
H.~H. Kim, S.~Shuzhi~Yu, and C.~Tomasi.
\newblock Joint detection of motion boundaries and occlusions.
\newblock In {\em British Machine Vision Conference (BMVC)}, November 2021.

\bibitem{Klappstein2009}
J.~Klappstein, T.~Vaudrey, C.~Rabe, A.~Wedel, and R.~Klette.
\newblock Moving object segmentation using optical flow and depth information.
\newblock In T.~Wada, F.~Huang, and S.~Lin, editors, {\em Advances in Image and
  Video Technology}, pages 611--623, Berlin, Heidelberg, 2009. Springer Berlin
  Heidelberg.

\bibitem{lei2018boundary}
P.~Lei, F.~Li, and S.~Todorovic.
\newblock Boundary flow: A siamese network that predicts boundary motion
  without training on motion.
\newblock In {\em Proceedings of the IEEE Conference on Computer Vision and
  Pattern Recognition}, pages 3282--3290, 2018.

\bibitem{arflowl2020liu}
L.~Liu, J.~Zhang, R.~He, Y.~Liu, Y.~Wang, Y.~Tai, D.~Luo, C.~Wang, J.~Li, and
  F.~Huang.
\newblock Learning by analogy: Reliable supervision from transformations for
  unsupervised optical flow estimation.
\newblock In {\em Proceedings of the IEEE/CVF Conference on Computer Vision and
  Pattern Recognition}, pages 6489--6498, 2020.

\bibitem{liu2019selflow}
P.~Liu, M.~Lyu, I.~King, and J.~Xu.
\newblock Selflow: Self-supervised learning of optical flow.
\newblock In {\em Proceedings of the IEEE Conference on Computer Vision and
  Pattern Recognition}, pages 4571--4580, 2019.

\bibitem{Lucas1981AnII}
B.~D. Lucas and T.~Kanade.
\newblock An iterative image registration technique with an application to
  stereo vision.
\newblock In {\em International Joint Conference on Artificial Intelligence
  (IJCAI)}, 1981.

\bibitem{maninis18osvoss}
K.-K. Maninis, S.~Caelles, Y.~Chen, J.~Pont-Tuset, L.~Leal-Taix\'e, D.~Cremers,
  and L.~{Van Gool}.
\newblock Video object segmentation without temporal information.
\newblock {\em IEEE Transactions on Pattern Analysis and Machine Intelligence
  (TPAMI)}, 2018.

\bibitem{bsds}
D.~Martin, C.~Fowlkes, D.~Tal, and J.~Malik.
\newblock A database of human segmented natural images and its application to
  evaluating segmentation algorithms and measuring ecological statistics.
\newblock In {\em International Conference on Computer Vision}, volume~2, pages
  416--423, July 2001.

\bibitem{meister2018unflow}
S.~Meister, J.~Hur, and S.~Roth.
\newblock Unflow: Unsupervised learning of optical flow with a bidirectional
  census loss.
\newblock In {\em Proceedings of the AAAI Conference on Artificial
  Intelligence}, 2018.

\bibitem{Menze2015ISA}
M.~Menze, C.~Heipke, and A.~Geiger.
\newblock Joint 3d estimation of vehicles and scene flow.
\newblock In {\em ISPRS Workshop on Image Sequence Analysis (ISA)}, 2015.

\bibitem{Menze2018JPRS}
M.~Menze, C.~Heipke, and A.~Geiger.
\newblock Object scene flow.
\newblock {\em ISPRS Journal of Photogrammetry and Remote Sensing (JPRS)},
  2018.

\bibitem{Meyer2018DeepVC}
S.~Meyer, V.~Cornill{\`e}re, A.~Djelouah, C.~Schroers, and M.~H. Gross.
\newblock Deep video color propagation.
\newblock In {\em British Machine Vision Conference (BMVC)}, 2018.

\bibitem{Narayana2013CoherentMS}
M.~Narayana, A.~Hanson, and E.~Learned-Miller.
\newblock Coherent motion segmentation in moving camera videos using optical
  flow orientations.
\newblock {\em 2013 IEEE International Conference on Computer Vision}, pages
  1577--1584, 2013.

\bibitem{park2021ABME}
J.~Park, C.~Lee, and C.-S. Kim.
\newblock Asymmetric bilateral motion estimation for video frame interpolation.
\newblock In {\em International Conference on Computer Vision}, 2021.

\bibitem{Ranjan2017OpticalFE}
A.~Ranjan and M.~J. Black.
\newblock Optical flow estimation using a spatial pyramid network.
\newblock {\em 2017 IEEE Conference on Computer Vision and Pattern Recognition
  (CVPR)}, pages 2720--2729, 2017.

\bibitem{ren2017unsupervised}
Z.~Ren, J.~Yan, B.~Ni, B.~Liu, X.~Yang, and H.~Zha.
\newblock Unsupervised deep learning for optical flow estimation.
\newblock In {\em Proceedings of the AAAI Conference on Artificial
  Intelligence}, 2017.

\bibitem{Sevilla-Lara2019}
L.~Sevilla-Lara, Y.~Liao, F.~G{\"u}ney, V.~Jampani, A.~Geiger, and M.~J. Black.
\newblock On the integration of optical flow and action recognition.
\newblock In T.~Brox, A.~Bruhn, and M.~Fritz, editors, {\em Pattern
  Recognition}, pages 281--297, Cham, 2019. Springer International Publishing.

\bibitem{xsoria2020dexined}
X.~Soria, E.~Riba, and A.~Sappa.
\newblock Dense extreme inception network: Towards a robust cnn model for edge
  detection.
\newblock In {\em 2020 IEEE Winter Conference on Applications of Computer
  Vision (WACV)}, pages 1912--1921, Los Alamitos, CA, USA, mar 2020. IEEE
  Computer Society.

\bibitem{stone2021smurf}
A.~Stone, D.~Maurer, A.~Ayvaci, A.~Angelova, and R.~Jonschkowski.
\newblock Smurf: Self-teaching multi-frame unsupervised raft with full-image
  warping.
\newblock In {\em Proceedings of the IEEE/CVF Conference on Computer Vision and
  Pattern Recognition}, pages 3887--3896, 2021.

\bibitem{sui2022craft}
X.~Sui, S.~Li, X.~Geng, Y.~Wu, X.~Xu, Y.~Liu, R.~Goh, and H.~Zhu.
\newblock Craft: Cross-attentional flow transformer for robust optical flow.
\newblock In {\em Proceedings of the IEEE/CVF Conference on Computer Vision and
  Pattern Recognition}, pages 17602--17611, 2022.

\bibitem{PWC}
D.~Sun, X.~Yang, M.-Y. Liu, and J.~Kautz.
\newblock Pwc-net: Cnns for optical flow using pyramid, warping, and cost
  volume.
\newblock In {\em Conference on Computer Vision and Pattern Recognition}, 2018.

\bibitem{raft}
Z.~Teed and J.~Deng.
\newblock Raft: Recurrent all-pairs field transforms for optical flow.
\newblock In {\em European Conference on Computer Vision}, pages 402--419.
  Springer, 2020.

\bibitem{Vihlman_Visala_2020}
M.~Vihlman and A.~Visala.
\newblock Optical flow in deep visual tracking.
\newblock {\em Proceedings of the AAAI Conference on Artificial Intelligence},
  34(07):12112--12119, Apr. 2020.

\bibitem{wang2018occlusion}
Y.~Wang, Y.~Yang, Z.~Yang, L.~Zhao, P.~Wang, and W.~Xu.
\newblock Occlusion aware unsupervised learning of optical flow.
\newblock In {\em Proceedings of the IEEE Conference on Computer Vision and
  Pattern Recognition}, pages 4884--4893, 2018.

\bibitem{deepflow}
P.~Weinzaepfel, J.~Revaud, Z.~Harchaoui, and C.~Schmid.
\newblock {DeepFlow: Large displacement optical flow with deep matching}.
\newblock In {\em {IEEE Intenational Conference on Computer Vision}}, Sydney,
  Australia, Dec. 2013.

\bibitem{LDMB}
P.~Weinzaepfel, J.~Revaud, Z.~Harchaoui, and C.~Schmid.
\newblock {Learning to Detect Motion Boundaries}.
\newblock In {\em {Conference on Computer Vision and Pattern Recognition}},
  Boston, United States, 2015.

\bibitem{xu2022gmflow}
H.~Xu, J.~Zhang, J.~Cai, H.~Rezatofighi, and D.~Tao.
\newblock Gmflow: Learning optical flow via global matching.
\newblock In {\em Proceedings of the IEEE/CVF Conference on Computer Vision and
  Pattern Recognition}, pages 8121--8130, 2022.

\bibitem{yuan2022optical}
S.~Yuan, X.~Sun, H.~Kim, S.~Yu, and C.~Tomasi.
\newblock Optical flow training under limited label budget via active learning.
\newblock {\em arXiv preprint arXiv:2203.05053}, 2022.

\end{thebibliography}

\end{document}